\documentclass[letterpaper, 10 pt, conference]{ieeeconf}  

\usepackage{times}
\usepackage[numbers]{natbib}
\usepackage{multicol}
\usepackage[bookmarks=true]{hyperref}

\usepackage{lipsum}
\usepackage{comment}
\usepackage{balance}
\usepackage{graphicx}
\usepackage[table]{xcolor}
\usepackage{caption}
\usepackage{listings}
\usepackage{xcolor}

\definecolor{codegreen}{rgb}{0,0.6,0}
\definecolor{codegray}{rgb}{0.5,0.5,0.5}
\definecolor{codepurple}{rgb}{0.58,0,0.82}
\definecolor{backcolour}{rgb}{0.95,0.95,0.92}

\lstdefinestyle{mystyle}{
    backgroundcolor=\color{backcolour},
    commentstyle=\color{codegreen},
    keywordstyle=\color{magenta},
    numberstyle=\tiny\color{codegray},
    stringstyle=\color{codepurple},
    basicstyle=\ttfamily,
    breaklines=true,
    captionpos=b,
    keepspaces=true,
    numbers=left,
    numbersep=5pt,
    showstringspaces=false,
    tabsize=2
}
\usepackage{minted}
\usepackage{xcolor}
\definecolor{bg}{HTML}{f2f2ea}
\usepackage{float} 

\newcommand{\cmark}{\ding{51}} 
\newcommand{\gmark}{\textcolor{LimeGreen}{\ding{51}}} 
\usepackage{pifont}

\usepackage{colortbl}
\usepackage{amsmath}
\usepackage{mathtools}
\usepackage{amsfonts}
\usepackage{svg}
\usepackage{cuted}    
\usepackage{booktabs}
\usepackage{makecell}
\usepackage[dvipsnames]{xcolor}

\usepackage{seqsplit}

\IEEEoverridecommandlockouts                              

\usepackage{tabularx} 

\title{\Large \bf
Subsecond 3D Mesh Generation for Robot Manipulation
}

\author{\authorblockN{Qian Wang, Omar Abdellall$^{*}$\thanks{* Equal contribution}, Tony Gao$^{*}$, Xiatao Sun, Daniel Rakita}
\authorblockA{Department of Computer Science, 
Yale University}
\authorblockA{\{peter.wang.qw262, omar.abdellall, t.gao, xiatao.sun, daniel.rakita\}@yale.edu}
\thanks{This work was supported by Office of Naval Research award N00014-24-1-2124}}    
\begin{document}

\maketitle
\thispagestyle{empty}
\pagestyle{empty}

\begin{strip}
  \centering
  \includegraphics[width=0.99\textwidth]{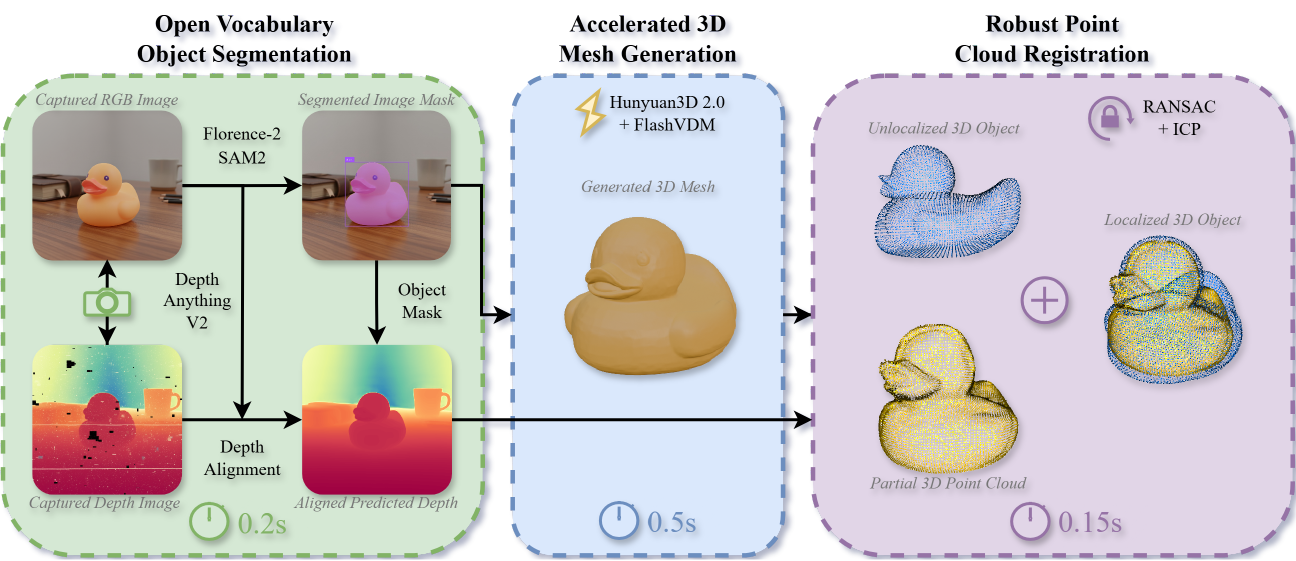}
  \captionof{figure}{Our system for sub-second 3D mesh generation from RGB-D input. The system combines three stages: (1) Open-vocabulary segmentation using Florence-2 and SAM2 with depth enhancement via Depth Anything v2 (0.2s), (2) Accelerated mesh generation using FlashVDM-distilled Hunyuan3D 2.0 (0.5s), and (3) Object registration via RANSAC and ICP to align the mesh with observed point cloud (0.15s). The 0.85s total runtime marks a critical step toward real-time robotic applications.}
  \label{fig:teaser}
  \vspace{-0.5cm}
\end{strip}
\begin{abstract}

3D meshes are a fundamental representation widely used in computer science and engineering. In robotics, they are particularly valuable because they capture objects in a form that aligns directly with how robots interact with the physical world, enabling core capabilities such as predicting stable grasps, detecting collisions, and simulating dynamics. Although automatic 3D mesh generation methods have shown promising progress in recent years, potentially offering a path toward real-time robot perception, two critical challenges remain. First, generating high-fidelity meshes is prohibitively slow for real-time use, often requiring tens of seconds per object. Second, mesh generation by itself is insufficient. In robotics, a mesh must be \textit{contextually grounded}, i.e., correctly \textit{segmented} from the scene and \textit{registered} with the proper scale and pose. Additionally, unless these contextual grounding steps remain efficient, they simply introduce new bottlenecks. In this work, we introduce an end-to-end system that addresses these challenges, producing a high-quality, contextually grounded 3D mesh from a single RGB-D image in under one second. Our pipeline integrates open-vocabulary object segmentation, accelerated diffusion-based mesh generation, and robust point cloud registration, each optimized for both speed and accuracy. We demonstrate its effectiveness in a real-world manipulation task, showing that it enables meshes to be used as a practical, on-demand representation for robotics perception and planning.

\end{abstract}

\section{Introduction}
\label{sec:introduction}

3D meshes are a fundamental representation in computer science and engineering. Defined by vertices, edges, and faces, they encode explicit object geometry and topology in a form that supports direct physical reasoning~\cite{botsch2010polygon}. Unlike representations such as images, voxel grids, signed distance functions, or radiance fields, meshes provide an explicit, physics-ready blueprint of the world, enabling algorithms to operate directly on geometry and surface detail.

In robotics, a mesh aligns closely with how robots must interact with the world, supporting essential tasks such as predicting stable grasps, planning collision-free trajectories, and simulating contact dynamics. If objects in a robot’s environment could be represented as high-fidelity meshes, much of the complexity of perception would simplify into a representation that is easily usable by downstream planning and control strategies.

Despite the representational power of 3D meshes in robotics, translating this representation to real-world environments at scale is a significant challenge. Manually created meshes, as used in graphics and game development, are impractical for robotics: robots must operate in unstructured, ever-changing environments where new objects continually appear. Early robotic systems often circumvented this challenge by assuming a finite set of known objects, using fiducial markers like AprilTags \cite{wang2016apriltag} or database lookups of CAD models \cite{samarawickrama20246d}. Although effective in constrained scenarios, such approaches are fundamentally unscalable in open-world environments.

Recent advances in generative modeling have begun to close this gap, producing detailed 3D meshes directly from RGB images. However, two critical challenges remain. First, high-fidelity mesh generation is prohibitively slow for real-time use, often requiring tens of seconds per object. Second, mesh generation alone is not sufficient in robotics. In this context, a mesh must be \textit{contextually grounded}, i.e., correctly \textit{segmented} from the scene and \textit{registered} with the proper scale and pose. Compounding this challenge, unless these additional contextual grounding steps remain efficient, they simply introduce new bottlenecks.

In this paper, we introduce the first end-to-end system capable of generating a high-quality, contextually grounded (i.e., segmented and registered) 3D mesh from a single RGB-D image in under one second. Our pipeline combines three tightly integrated components: (1) open-vocabulary segmentation to identify and isolate objects of interest; (2) an accelerated diffusion-based model to produce high-fidelity meshes; and (3) robust point cloud registration to recover the correct scale and pose within the scene. 

The mesh generation stage is based on Hunyuan3D 2.0 \cite{zhao2025hunyuan3d}, but, to overcome its prohibitive latency for robotics, we incorporate FlashVDM \cite{lai2025unleashing}, which applies progressive flow distillation to reduce the number of diffusion steps from dozens to as few as three. To further accelerate inference, we adopt hierarchical SDF decoding with adaptive key–value selection, reducing volumetric decoding costs by over 90\%. Together, these techniques deliver generative fidelity comparable to full diffusion pipelines while reducing runtime from tens of seconds to well under one second. By unifying these advances into a single low-latency pipeline, our system transforms mesh acquisition from a slow, offline process into a practical, on-demand tool toward real-time robotics.

To demonstrate the efficacy of our system, we conduct an evaluation with three sub-experiments (\S\ref{sec:evaluation}). First, we analyze the runtime of each component to confirm sub-second performance and identify bottlenecks. Second, we perform ablation studies comparing alternative designs for segmentation, generation, and registration, highlighting the trade-offs between speed and accuracy. Finally, we demonstrate the system in a real-world manipulation task, showing that it enables reliable on-demand grasping and placement using generated meshes. These results collectively establish that fast, contextually grounded mesh generation is not only feasible but also effective for real-time robotic interaction. We conclude with a discussion of the limitations and implications of our work.
\section{Related Works}
\label{sec:related_works}

Our proposed pipeline combines modules for object detection and segmentation, single-image 3D mesh generation, and object localization. This section reviews each of these areas and contextualizes our methodological choices.

\begin{figure}[]
    \centering
    \includegraphics[width=0.35\textwidth]{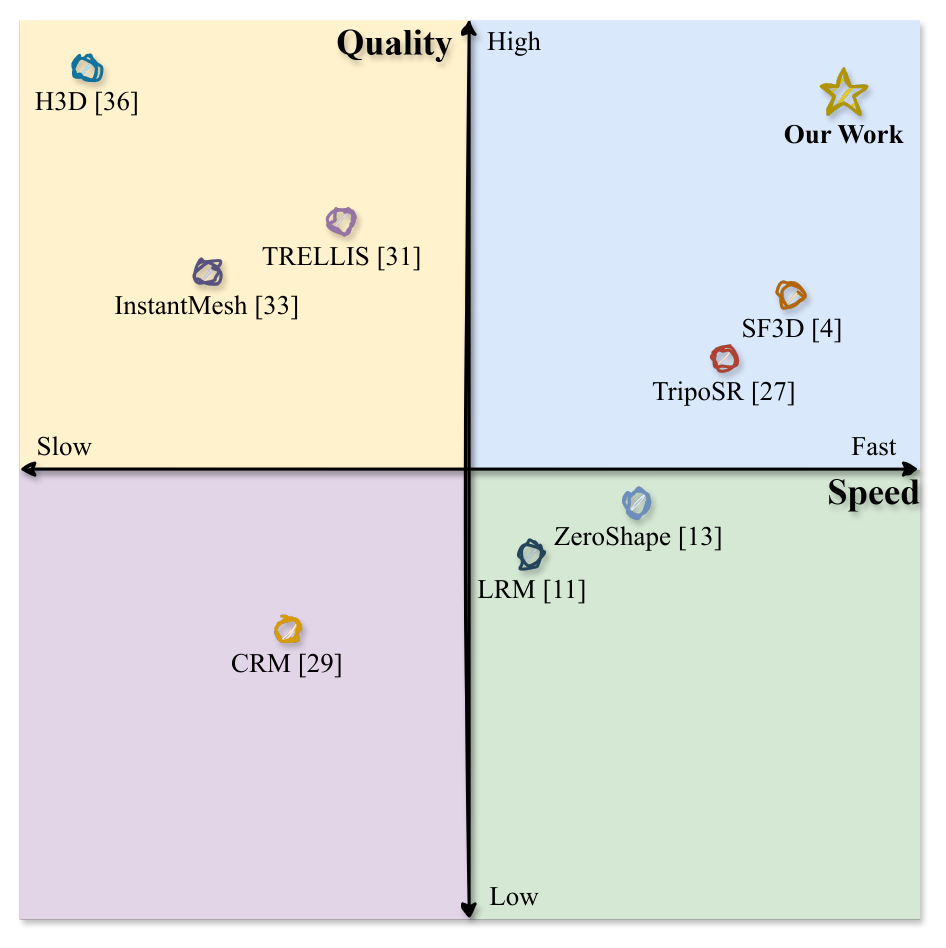}
    \caption{The Speed vs. Quality Trade-off in Single-Image 3D Generation. Existing methods force a choice between fast but lower-quality reconstruction and high-quality but slow generation. Our system is designed to achieve both, breaking a key barrier for real-time robotics.}
    \label{fig:tradeoff}
    \vspace{-0.65cm}
\end{figure}

\subsection{Single-Image 3D Mesh Generation}

Generating a 3D object from a single 2D image with both quality and speed is a long-standing challenge, defined by a fundamental tradeoff between inference speed and geometric quality, as illustrated in Fig.~\ref{fig:tradeoff}. Early works such as LRM \cite{honglrm}, CRM \cite{wang2024crm}, and InstantMesh \cite{xu2024instantmesh} established foundational model architectures but suffered from either slow inference or low geometric fidelity.

More recently, large-scale generative models have reached a new level of quality. Models like Hunyuan3D 2.0 (H3D)~\cite{zhao2025hunyuan3d} can generate high-resolution, textured 3D assets, while others like TRELLIS \cite{xiang2024structured} leverage structured priors to improve geometric consistency. However, the primary drawback of such models is their slow generation speed. Models like ZeroShape \cite{huang2024zeroshape}, TripoSR \cite{tochilkin2024triposr}, SF3D \cite{boss2025sf3d}, and \cite{sun2025hybrid} focused on improving efficiency, but this speed often came at the cost of reduced quality and limited generalization to novel object categories. To break this quality-speed trade-off, our system builds upon H3D's high-fidelity generation principles but applies two key modifications for acceleration. First, we integrate FlashVDM \cite{lai2025unleashing} to speed up the generation process. Second, we deliberately omit texture generation to focus solely on fast, high-quality shape reconstruction--a design choice motivated by the observation that most robotic tasks (grasping, collision avoidance, dynamics simulation) primarily require geometric rather than visual information.

\subsection{Open-Vocabulary Object Segmentation}

To generate a 3D mesh of a specific object, it must first be precisely segmented from the background. The field of universal image segmentation has produced powerful models like OneFormer \cite{jain2023oneformer} and Mask2Former \cite{cheng2022masked}. Our work employs the Segment Anything Model 2 (SAM2) \cite{ravi2024sam}, which currently represents the state-of-the-art, offering strong zero-shot generalization for both images and videos. However, SAM2 requires an initial point, bounding box, or mask to identify the target object. In a robotic system operating autonomously, this prompting must be automated through object detection.

Traditional object detection has seen rapid progress, from real-time architectures like YOLOv12 \cite{tian2025yolov12} to Transformer-based models like DINOv2 \cite{oquab2023dinov2} and D-FINE \cite{pengd}, which have achieved higher accuracy on standard benchmarks. While effective, these models are typically trained on a fixed set of classes, fundamentally limiting their flexibility in open-world robotic scenarios where novel objects frequently appear. The solution lies in Vision-Language Models (VLMs) that can detect objects from free-form text prompts. These models exist on a spectrum of scale and specialization. At the high end, massive generalist models like Flamingo \cite{alayrac2022flamingo} (80B parameters) and PaLM-E \cite{driess2023palm} (562B parameters) demonstrate impressive multi-modal reasoning but are computationally prohibitive for real-time pipelines. In contrast, more specialized and efficient models have been developed for targeted vision tasks. For instance, Florence-2 \cite{xiao2024florence} (0.23B parameters) excels at grounding text in images for which the small size makes it ideal for a performance-critical system.

\subsection{Object Localization and Spatial Grounding}

Once a 3D mesh is generated, it must be localized within the robot's coordinate frame, recovering both the correct scale and 6D pose relative to the observed scene. Recent literature has converged on render-and-refine methods \cite{ornek2024foundpose} for this task. FoundationPose \cite{wen2024foundationpose}, for instance, performs 6D pose estimation by iteratively aligning rendered views of an object with sensor data, which has been deployed in multiple robotic systems \cite{agarwal2024scenecomplete, sun2025prism, sun2025hybrid}. However, these methods fundamentally rely on photorealistic rendering, a process which requires a textured mesh. As our pipeline deliberately forgoes texture generation to maximize speed, these render-based approaches are incompatible with our output. 

Therefore, we employ direct geometric alignment via point cloud registration. We employ RANSAC \cite{fischler1981random}, a fast and robust method for model fitting in the presence of outliers. We also investigated more recent registration algorithms, such as TEASER++ \cite{yang2020teaser} for certifiable optimality and BUFFER-X \cite{seo2025buffer} for zero-shot generalization, our empirical evaluations (\S\ref{sec:evaluation}) demonstrate that a well-tuned RANSAC implementation provides an effective combination of speed, accuracy, and robustness for our specific task.

\section{Method}
\label{sec:technical_details}

Our system generates a registered, high-fidelity 3D mesh from a single RGB-D image in under one second. As illustrated in Fig.~\ref{fig:teaser}, the system consists of three sequential stages: open-vocabulary segmentation, accelerated mesh generation, and robust point cloud registration.


\subsection{Open Vocabulary Image Segmentation}
The first stage of the pipeline identifies and segments objects of interest from the input image. To enable flexible, open-vocabulary segmentation, we leverage the VLM \textit{Florence-2} \cite{xiao2024florence}, which supports multiple segmentation paradigms. In robotics contexts, two functionalities of Florence-2 are particularly relevant: (1) prompt-free, generic object-level segmentation and (2) text-guided object or category-level segmentation. In this work, we primarily use the text-guided mode. Given a prompt specifying target object classes (e.g., ``fruits'') or regions (e.g., ``object on the table''), Florence-2 generates labeled bounding boxes for each instance. These serve as coarse outlines, which are then refined using SAM2 \cite{ravi2024sam}, yielding precise, pixel-level instance masks. Finally, these masks are then used to crop both the RGB image and its associated depth map, producing segmented object images with transparent backgrounds and corresponding partial depth maps.


\subsection{Processing Depth Input}

\begin{figure}[t]
    \centering
    \includegraphics[width=\linewidth]{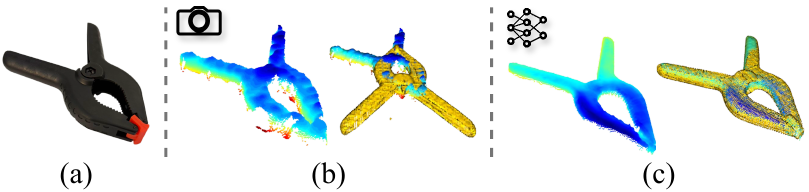}
    \caption{Effect of Depth Map Quality on Registration. (a) The source object. (b) A noisy point cloud from the raw depth camera feed results in a failed registration (misaligned overlap). (c) The cleaner, more coherent point cloud predicted by DAv2 enables a successful and accurate registration.}
    \label{fig:pointcloud}
    \vspace{-0.5cm}
\end{figure}

While the RGB image drives mesh generation, accurate depth is essential for scaling and localizing the result. Raw measurements from consumer-grade depth cameras suffer from noise, missing, regions, and artifacts on reflective or textureless surfaces (Fig.~\ref{fig:pointcloud}b). To address this issue, we employ Depth Anything v2 (DAv2) \cite{yang2024depth}, a monocular depth estimation model that predicts clean, geometrically consistent depth maps from RGB images.

However, monocular predictions lack metric scale. We resolve this issue through median-based alignment:
\begin{equation}
s = \frac{\text{median}(D_{\text{sensor}} \cap M)}{\text{median}(D_{\text{DAv2}} \cap M)}, \quad D_{\text{metric}} = s \cdot D_{\text{DAv2}}
\end{equation}
where $M$ is the object mask, and the median is computed only over valid (non-zero) depth values. We use median rather than mean as it is robust to the depth outliers common in consumer sensors. This approach combines the geometric consistency of learned depth with the metric accuracy of physical measurements, providing a reliable foundation for downstream stages in the pipeline.



\subsection{Vecset Diffusion Model for 3D Mesh Generation}

For 3D shape generation, we build on \textit{Hunyuan3D 2.0 (H3D)} \cite{zhao2025hunyuan3d}, a state-of-the-art flow-based diffusion transformer designed for high-fidelity mesh synthesis from a single image. The framework consists of two main components: a Shape VAE that compresses 3D meshes into a structured latent space, and a conditional diffusion model that generates these latents from a single image.

The Shape VAE consists of an encoder $\mathcal{E}$ and a decoder $\mathcal{D}$. The encoder maps a 3D mesh $M$, represented as a Signed Distance Function (SDF), into a compact set of $N$ latent vectors $Z = \{z_i\}_{i=1}^N \in \mathbb{R}^{N \times D}$, where $Z = \mathcal{E}(M)$. The decoder reconstructs the SDF from these latents: $\widehat{\text{SDF}} = \mathcal{D}(Z)$. The VAE is trained to minimize a reconstruction loss, typically a weighted L2 norm on the SDF values, combined with a KL-divergence regularization term on the latent distribution.

\begin{figure}[t]
    \centering
    \includegraphics[width=\linewidth]{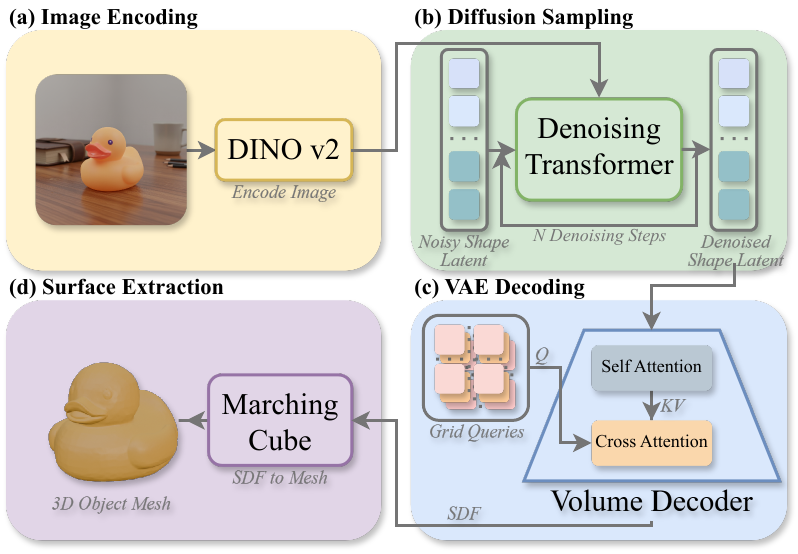}
    \caption{Standard Vecset Diffusion Model (VDM) pipeline for 3D mesh generation. The process consists of four stages: (a) image encoding using DINOv2 to extract visual features, (b) iterative diffusion sampling requiring N denoising steps (typically 50+), (c) VAE decoding to convert latent vectors to a dense SDF volume, and (d) mesh extraction via marching cubes. The two primary bottlenecks preventing real-time performance are the iterative diffusion process (b) and volumetric decoding (c).}
    \label{fig:diffusion}
    \vspace{-0.5cm}
\end{figure}

The generative model is a flow-based diffusion transformer, conditioned on image features $c$ extracted by a DINOv2 \cite{oquab2023dinov2} backbone (seen in Fig. \ref{fig:diffusion}(a)). It learns to transform a simple prior distribution $p_0$ (e.g., a standard normal distribution $\mathcal{N}(0, I)$) into the complex data distribution of shape latents $p_1(Z|c)$. This transformation is learned via a time-dependent vector field $v_\theta(Z_t, t, c)$ that follows the probability path from noise to data, governed by the ordinary differential equation (ODE):
\begin{equation}
    \frac{dZ_t}{dt} = v_t(Z_t, c),
\end{equation}
where $Z_t$ is the latent representation at time $t \in [0, 1]$. A flow-matching objective is used to train the model, which provides a stable and efficient regression target for the vector field. The loss function is formulated as:
\begin{equation}
    \mathcal{L}_{\text{FM}} = \mathbb{E}_{t, p_1(Z_1|c), p_0(Z_0)} \left[ \left\| v_\theta(Z_t, t, c) - (Z_1 - Z_0) \right\|^2_2 \right]
\end{equation}
where $Z_1$ is a ground-truth latent from the data distribution, $Z_0$ is a sample from the noise distribution, and the term $(Z_1 - Z_0)$ is the vector field of a linear probability path.


\subsection{Accelerating 3D Generation}

The standard diffusion model inference process, outlined above, requires solving an ODE over many steps, leading to runtimes far too slow for robotics (often tens of seconds per mesh). In this section, we present our approach for overcoming this limitation.  

\subsubsection{Vector-Set Diffusion Acceleration}

Diffusion sampling (Fig.~\ref{fig:diffusion}b) often requires dozens of sequential denoising steps, making it a dominant runtime cost. We employ \textit{Progressive Flow Distillation}, introduced in FlashVDM (FVDM) \cite{lai2025unleashing}, to distill the multi-step teacher model into a student capable of producing comparable results in as few as 3 steps where the original H3D model requires at least 50 steps.  

\subsubsection{Volume Decoding Acceleration}
The second bottleneck arises in decoding (Fig.~\ref{fig:diffusion}(c)), where shape latents must be converted into a dense SDF volume (e.g., $384^3$ voxels). A naive implementation requires millions of costly cross-attention evaluations. To address this, two complementary acceleration techniques from FVDM are adopted: 1) Hierarchical Volume Decoding: Instead of decoding the entire high-resolution grid, a coarse grid is decoded first, and only the voxels near the object's surface are refined. This reduces the number of queries by over 90\%. 2) Adaptive KV Selection: The locality of latent tokens is exploited by using a lightweight attention pass to pre-select only the most relevant key-value pairs for each query point, cutting the computational cost of attention by over 30\%.

\begin{figure*}[t]
    \centering
    \includegraphics[width=0.80\linewidth]{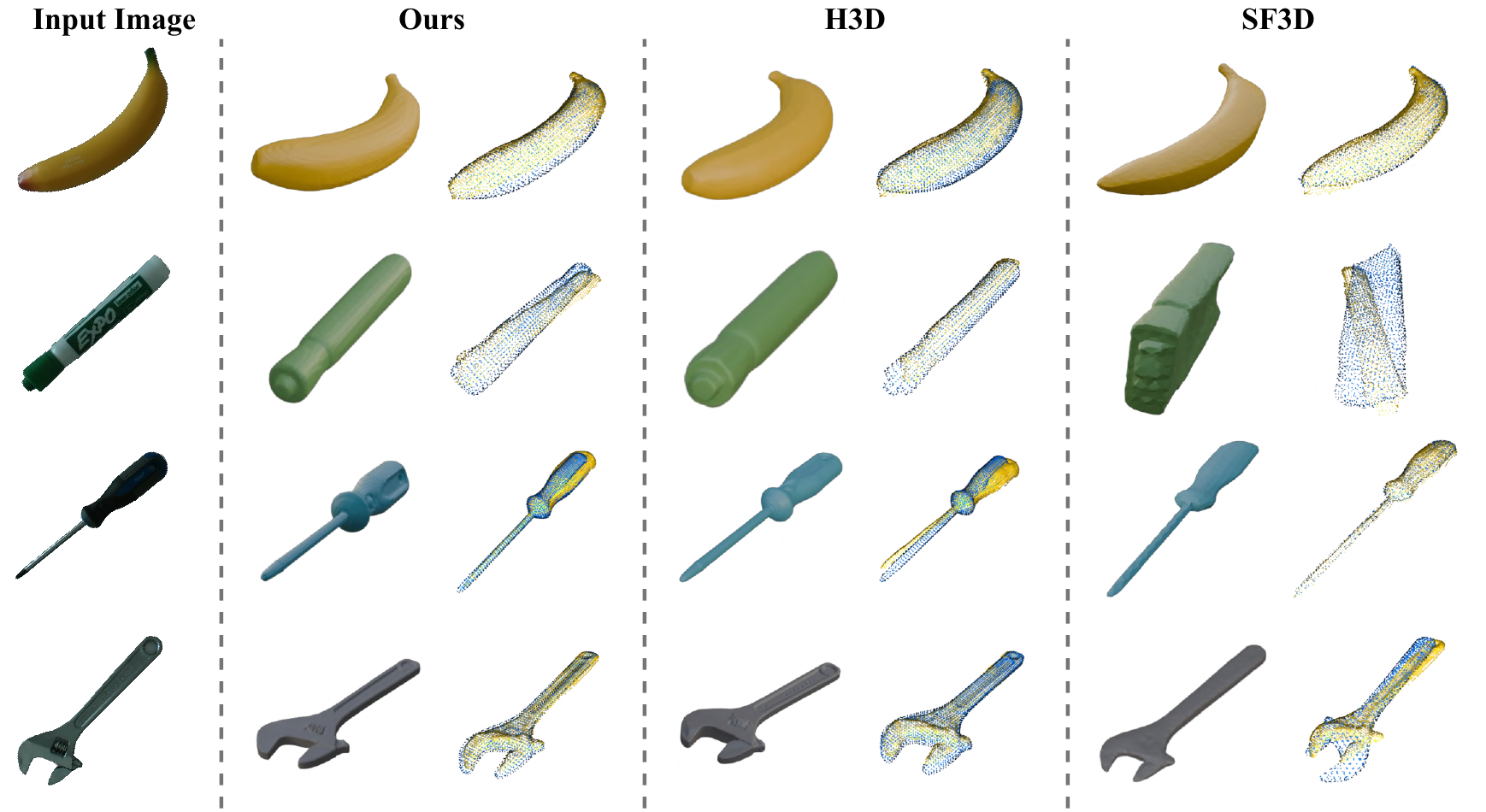}
    \caption{Qualitative Comparison of Generated Meshes and Registration Results. Our method (left) achieves high geometric quality nearly identical to the slow H3D baseline (middle), while the fast SF3D baseline (right) produces significant artifacts.}
    \label{fig:qualitative_results}
    \vspace{-5pt}
\end{figure*}

\subsection{Object Registration}

The final stage aligns the canonically generated mesh with the object's real-world observation. We formulate this process as a rigid point cloud registration problem. Let the source point cloud derived from our generated mesh be $P_S = \{\mathbf{p}_i^S\}_{i=1}^{N_S} \subset \mathbb{R}^3$ and the target point cloud from the depth sensor be $P_T = \{\mathbf{p}_j^T\}_{j=1}^{N_T} \subset \mathbb{R}^3$. The goal is to find an optimal scale $s \in \mathbb{R}^+$, rotation $R \in SO(3)$, and translation $\mathbf{t} \in \mathbb{R}^3$ that best align $P_S$ with $P_T$.

Our registration procedure proceeds as follows:

\subsubsection{Initial Scaling} We first uniformly scale $P_S$ to match the bounding box diagonal of $P_T$.

\subsubsection{Feature Extraction and Matching} We compute Fast Point Feature Histograms (FPFH) \cite{rusu2009fast} for both point clouds. For each point $\mathbf{p}_i$, FPFH computes a $k$-dimensional feature descriptor $\mathbf{f}(\mathbf{p}_i) \in \mathbb{R}^k$ that encodes the local geometry of its neighborhood. We then establish a set of putative correspondences $\mathcal{C} = \{(\mathbf{p}_i^S, \mathbf{p}_i^T)\}_{i=1}^N$ by finding mutual nearest neighbors in this high-dimensional feature space.

\subsubsection{Robust Pose Estimation with RANSAC} The correspondence set $\mathcal{C}$ from FPFH matching contains a significant fraction of outliers. To robustly estimate the transformation, we employ RANSAC (Random Sample Consensus)~\cite{fischler1981random}. The algorithm iteratively performs the following steps: 1) A minimal subset of correspondences (e.g., 3 pairs) is randomly selected from $\mathcal{C}$. 2) A rigid transformation $(R, \mathbf{t})$ is computed from this minimal sample. 3) The hypothesis is tested against all other correspondences in $\mathcal{C}$. Correspondences that agree with the transformation (i.e., where $\| sR\mathbf{p}_i^S + \mathbf{t} - \mathbf{p}_i^T \|_2$ is below a distance threshold) are counted as inliers. 4) Repeat: This process is repeated for a fixed number of iterations, and the transformation with the largest inlier set is chosen as the best hypothesis. This iterative process discards outliers and provides a robust initial alignment for the two point clouds.

\subsubsection{Refinement with ICP} The estimate from RANSAC is used to initialize an Iterative Closest Point (ICP) \cite{besl1992method} algorithm for fine-grained refinement. ICP iteratively performs two steps until convergence: (1) For each transformed source point $sR\mathbf{p}_i^S + \mathbf{t}$, find its closest neighbor in the target cloud, $\mathbf{p}_i^{T'} \in P_T$; and (2) Given correspondences $\{(x_i, y_i)\}_{i=1}^N$ with $x_i = sR\mathbf{p}_i^S+\mathbf{t}$ fixed from the previous iteration, solve
\begin{equation}
    \min_{R',\,\mathbf{t'}} \sum_{i=1}^{N} \| R' x_i + \mathbf{t'} - y_i \|_2^2 \quad \text{s.t. } R' \in SO(3).
\end{equation}

Let $\bar{x}=\frac{1}{N}\sum_i x_i$, $\bar{y}=\frac{1}{N}\sum_i y_i$, and $x_i' = x_i-\bar{x}$, $y_i' = y_i-\bar{y}$. Form $H=\sum_i x_i' {y_i'}^\top$ and compute its SVD $H=U\Sigma V^\top$. The optimal update is
\begin{align}
    R'        &= V \,\mathrm{diag}(1,1,\det(VU^\top))\, U^\top, \\
    \mathbf{t'} &= \bar{y} - R'\bar{x},
\end{align}
        
the standard orthogonal Procrustes solution~\cite{arun1987least,horn1987closed}. 

This process yields an accurate and robust alignment of the generated mesh in the scene, ready for downstream manipulation tasks.

\section{Evaluation}
\label{sec:evaluation}

We evaluate our system on three fronts: (1) component latency and bottlenecks, (2) comparisons against alternative designs, and (3) a real-world robotic manipulation task. Experiments were run on a workstation with AMD 7960X CPU, 256GB RAM, and a RTX 5000 Ada GPU.

\subsection{Component-wise Time Breakdown}

We measure the runtime of each stage's sub components, averaged over 10 runs on 25 YCB objects \cite{calli2017yale}. As shown in Table~\ref{tab:time_breakdown}, the entire pipeline, from a single RGB image to a fully registered mesh, executes in 824 ms on average. Mesh generation dominates the runtime at 500ms (60.7\%), with the distilled diffusion process being the primary contributor. The segmentation and registration stages are well-optimized at 184ms and 140ms respectively, with no single component creating a prohibitive bottleneck.

The relatively high variance in registration time ($\pm$61ms) stems from object complexity—simple geometric primitives converge quickly while complex, asymmetric objects require more RANSAC iterations. Despite this variance, even worst-case scenarios remain under our 1-second target.

\begin{table}[h]
\centering
\caption{Component Runtime Breakdown. Average runtime and percentage of total pipeline time for each subcomponent, measured over 250 trials.}
\label{tab:time_breakdown}
\begin{tabular}{lcc}
\toprule
\textbf{Pipeline Stage} & \textbf{Time (ms)} & \textbf{Time (\%)} \\
\midrule
\rowcolor{gray!20}
\textbf{Open-Vocabulary Segmentation} & \textbf{184 $\pm$ 7.3} & \textbf{22.3\%} \\
\quad \textit{- Florence-2 Detection} & \textit{(55 $\pm$ 2.5)} & \textit{(6.7\%)} \\
\quad \textit{- SAM2 Mask Refinement} & \textit{(14 $\pm$ 0.3)} & \textit{(1.7\%)} \\
\quad \textit{- Depth Anything v2} & \textit{(93 $\pm$ 6.3)} & \textit{(11.3\%)} \\
\midrule
\rowcolor{gray!20}
\textbf{Accelerated 3D Mesh Generation} & \textbf{500 $\pm$ 37} & \textbf{60.7\%} \\
\quad \textit{- DINOv2 Image Encoding} & \textit{(10 $\pm$ 0.5)} & \textit{(1.2\%)} \\
\quad \textit{- Distilled Diffusion (3 steps)} & \textit{(225 $\pm$ 2)} & \textit{(27.3\%)} \\
\quad \textit{- Hierarchical SDF Decoding} & \textit{(104 $\pm$ 3.4)} & \textit{(12.6\%)} \\
\quad \textit{- Marching Cube (Res: 128)} & \textit{(43 $\pm$ 22)} & \textit{(5.2\%)} \\
\midrule
\rowcolor{gray!20}
\textbf{Object Registration} & \textbf{140 $\pm$ 61} & \textbf{17\%} \\
\quad \textit{- FPFH Feature Extraction} & \textit{(8 $\pm$ 4)} & \textit{(1\%)} \\
\quad \textit{- RANSAC Pose Estimation} & \textit{(79 $\pm$ 20)} & \textit{(9.6\%)} \\
\quad \textit{- ICP Refinement} & \textit{(57 $\pm$ 53)} & \textit{(7\%)} \\
\midrule
\textbf{Total End-to-End Time} & \textbf{824 $\pm$ 95} & \\
\bottomrule
\end{tabular}
\vspace{-0.4cm}
\end{table}

\begin{table*}[t!]
\centering
\caption{Ablation Study. Each row evaluates a complete pipeline, with checkmarks indicating the active module for each configuration. The results validate that our system (top row) achieves the optimal balance of speed and geometric accuracy.}
\label{tab:ablation_summary}
\begin{tabular}{c!{\color{lightgray}\vline}c!{\color{lightgray}\vline}c|c!{\color{lightgray}\vline}c!{\color{lightgray}\vline}c!{\color{lightgray}\vline}c|c!{\color{lightgray}\vline}c!{\color{lightgray}\vline}c||c!{\color{lightgray}\vline}c!{\color{lightgray}\vline}c!{\color{lightgray}\vline}c!{\color{lightgray}\vline}c}
\toprule
\multicolumn{3}{c|}{Depth Input} & \multicolumn{4}{c|}{Mesh Generation} & \multicolumn{3}{c||}{Registration} & \multicolumn{5}{c}{Metrics} \\
\midrule
\rotatebox{90}{\makecell[l]{DAv2 + \\Align (ours)}} & \rotatebox{90}{DAv2} & \rotatebox{90}{Depth Cam.} & \rotatebox{90}{\makecell[l]{H3D + \\FVDM (ours)}} & \rotatebox{90}{H3D} & \rotatebox{90}{TRELLIS} & \rotatebox{90}{SF3D} & \rotatebox{90}{\makecell[l]{RANSAC \\(ours)}} & \rotatebox{90}{TEASER++} & \rotatebox{90}{BUFFER-X} & \makecell{Component \\ time (s)} & \makecell{Component \\ GPU Peak \\ Memory (GB)} & \makecell{Total \\ time (s)} & \makecell{Mesh Chamfer \\ Distance (mm)} & F-Score (\%) \\
\midrule
\gmark & & & \cmark & & & & \cmark & & & \textbf{0.09 $\pm$ 0.01} & \textbf{1.96} & \textbf{0.83 $\pm$ 0.10} & \textbf{0.45 $\pm$ 0.90} & \textbf{89.9 $\pm$ 21.8} \\ 
\arrayrulecolor{lightgray}\hline
 & \gmark & & \cmark & & & & \cmark & & & 0.09 $\pm$ 0.01 & 1.96 & 0.83 $\pm$ 0.09 & 106 $\pm$ 51.1 & 0.12 $\pm$ 0.41 \\ 
\hline
 & & \gmark & \cmark & & & & \cmark & & & 0.03 $\pm$ 0.01 & N/A & 0.77 $\pm$ 0.11 & 0.62 $\pm$ 1.29 & 87.7 $\pm$ 23.6 \\ 
\hline
\cmark & & & \gmark & & & & \cmark & & & 0.50 $\pm$ 0.04 & 5.11 & \textit{= row1} & \textit{= row1} & \textit{= row1} \\ 
\hline
\cmark & & & & \gmark & & & \cmark & & & 30.1 $\pm$ 0.17 & 7.45 & 30.6 $\pm$ 0.27 & 0.34 $\pm$ 0.65 & 90.6 $\pm$ 19.5 \\ 
\hline
\cmark & & & & & \gmark & & \cmark & & & 4.41 $\pm$ 1.08 & 10.6 & 4.78 $\pm$ 1.12 & 44.2 $\pm$ 118 & 59.8 $\pm$ 34.2 \\ 
\hline
\cmark & & & & & & \gmark & \cmark & & & 0.53 $\pm$ 0.03 & 6.17 & 0.83 $\pm$ 0.07 & 0.79 $\pm$ 1.00 & 75.2 $\pm$ 28.4 \\ 
\hline
\cmark & & & \cmark & & & & \gmark & & & 0.14 $\pm$ 0.06 & N/A & \textit{= row1} & \textit{= row1} & \textit{= row1} \\ 
\hline
\cmark & & & \cmark & & & & & \gmark & & 0.08 $\pm$ 0.05 & N/A & 0.78 $\pm$ 0.09 & 0.79 $\pm$ 1.22 & 77.5 $\pm$ 25.0 \\ 
\hline
\cmark & & & \cmark & & & & & & \gmark & 0.28 $\pm$ 0.13 & 1.05 & 0.98 $\pm$ 0.10 & 118 $\pm$ 159 & 49.9 $\pm$ 43.8 \\ 
\arrayrulecolor{black}\bottomrule
\end{tabular}
\vspace{-0.5cm}
\end{table*}

\subsection{Ablation Studies}

\subsubsection{Procedure}
We systematically evaluate alternative components for each pipeline stage using 25 YCB objects. For each alternative, we replace only the specified component while keeping others fixed, enabling direct assessment of each design choice's impact on the complete system.

\subsubsection{Metrics}
In addition to average component time, component GPU peak memory, and total time, we evaluate the geometric quality of the final registered mesh against its ground-truth localized model using two widely adopted metrics from previous works~\cite{boss2025sf3d}: 1) Chamfer Distance (CD): This metric computes the average symmetric closest-point distance between two point clouds, providing a general measure of surface deviation. 2) F-Score: The harmonic mean of precision and recall, calculated at a 2cm threshold. This measures the percentage of the mesh surface that is accurately reconstructed. Furthermore, these two metrics together give a sense of how the object's pose is misaligned against ground truth.

\subsubsection{Results}
The results, summarized in Table \ref{tab:ablation_summary}, validate that our chosen components achieve the best overall balance of speed, accuracy, and efficiency.

Our DAv2 + alignment approach is critical for accurate registration. Using DAv2 alone fails catastrophically (106mm CD) due to lack of metric scale. Raw depth is faster but noisier, resulting in slightly degraded registration quality (0.62mm vs. 0.45mm CD).

Results from the mesh generation component validate our design choice to adopt FVDM's acceleration of H3D for robotic applications. The FVDM-accelerated H3D achieves quality nearly identical to the original (89.9\% vs. 90.6\% F-Score) while being 37× faster—a critical improvement that makes subsecond mesh generation feasible. SF3D matches our speed but sacrifices too much quality (75.2\% F-Score). TRELLIS is both slower and less accurate than ours.

For Registration, our classical FPFH + RANSAC + ICP (RANSAC) approach provides the most accurate and robust alignment. FPFH + TEASER++ + ICP (TEASER++) is marginally faster (80ms vs. 140ms) but less accurate. The learning-based BUFFER-X suffers from significant generalization errors, making it unsuitable for reliable deployment.

These quantitative findings are further supported by our qualitative results in Fig. \ref{fig:qualitative_results}, showing that our pipeline is a carefully optimized system, with each component chosen not only for individual performance but also for its contribution to the overall speed-accuracy balance.

\subsection{Real-world Online Pick-and-Place}

\begin{figure}[t!]
    \centering
    \includegraphics[width=0.8\columnwidth]{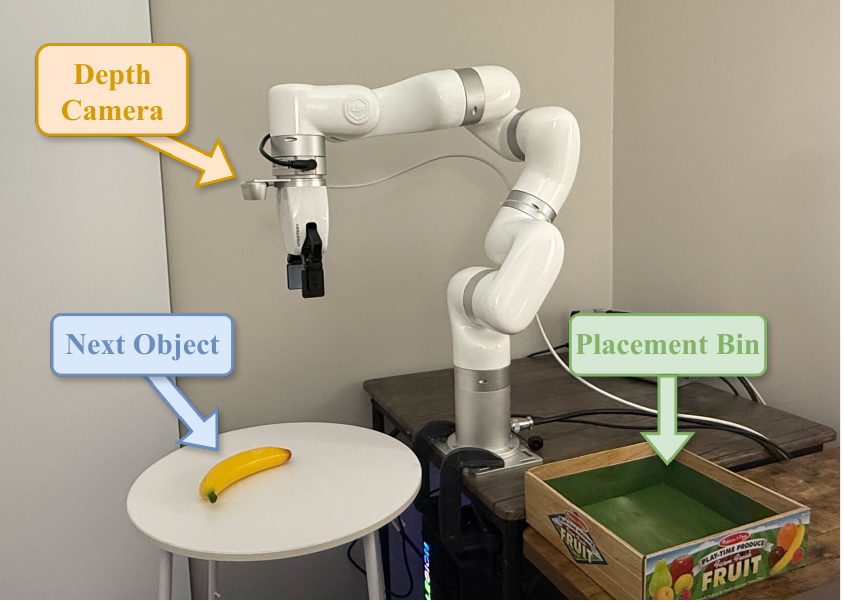}
    \caption{Real-world manipulation setup. The xArm7 robot uses our generated meshes to grasp and transfer unknown objects. The system processes each object in real-time without prior CAD models.}
    \label{fig:robot_setup}
    \vspace{-15pt}
\end{figure}

\subsubsection{Procedure}
We validate our system's practicality through a sequential pick-and-place task using a UFactory XArm7 robot (Fig.~\ref{fig:robot_setup}). The robot must perceive novel objects placed on a table, generate their 3D meshes, plan collision-free grasps, and transfer them to a target bin.

\subsubsection{Implementation Details} 
Once an object's mesh is generated and registered in the camera frame, its pose is transformed to the robot's base frame using a pre-calibrated extrinsic matrix. To find a stable grasp, we use a simple sampling-based planner sufficient for this demonstration; the method samples a set of end-effector poses by perturbing the translation and rotation around the object's centroid, retaining the first collision-free grasp found. Before execution, promising grasps are validated in a PyBullet simulation \cite{coumans2016pybullet} to filter out kinematically unreachable poses. Finally, a collision-free trajectory is computed using the RRT algorithm \cite{lavalle1998rapidly}, with the generated mesh used for collision checking.

\begin{table}[h]
\centering
\caption{Pick-and-Place Performance. Our system achieves both high reliability and fastest task completion.}
\label{tab:robot_demo}
\begin{tabular}{l|c|c}
\toprule
\textbf{System Configuration} & \textbf{Total Success Rate} & \textbf{Makespan (s)} \\
\midrule
\textbf{Ours} & \textbf{92\%} & \textbf{122} \\
SF3D & 60\% & 129 \\
H3D & 96\% & 416 \\
\bottomrule
\end{tabular}
\end{table}

\subsubsection{Metrics}
We evaluate sytem performance using two complementary metrics: 1) Total success rate: Percentage of objects successfully grasped, lifted, and placed in the target bin without drops or collisions. 2) Makespan: Total time to complete all 10 sequential pick-and-place operations, from initial perception to final placement.

\subsubsection{Results}
Table~\ref{tab:robot_demo} demonstrates our system's practical advantages. While H3D achieves marginally higher success (96\% vs. 92\%), its 30-second generation time makes the complete task take 3.4× longer (416s vs. 122s). This delay would be prohibitive in dynamic environments where objects frequently change. SF3D's poor mesh quality leads to cascading failures: inaccurate geometry causes registration errors, which result in grasps missing the actual object surface. Of its 10 failures, 7 were complete misses and 3 were unstable grasps that slipped during lifting.
\section{Discussion}

\label{sec:discussion}
Our work demonstrates that high-quality, contextually grounded 3D meshes can be generated from single RGB-D images in under one second, making them viable for real-time robotic applications. By integrating open-vocabulary segmentation, FlashVDM-accelerated mesh synthesis, and robust geometric registration, we achieve both the speed required for interactive systems and the geometric fidelity needed for manipulation tasks. Real-world experiments validate this balance—our system completed pick-and-place tasks in 122 seconds with 92\% success, significantly faster than prior methods while maintaining reliability.

\subsection{Limitations}
Several constraints bound the current system's applicability. First, we omit texture generation to maintain sub-second performance, limiting applications requiring appearance information for semantic reasoning or human interaction. Second, the system assumes full object visibility and fails under heavy occlusion or partial views, though our speed enables reactive regeneration as viewpoints change. Third, our mesh representation only handles rigid objects, excluding deformable materials common in many robotic tasks. Finally, while robust in controlled settings, deployment in cluttered, unstructured environments will require improvements in segmentation and depth estimation.

Despite these limitations, this work establishes that sub-second mesh generation is both technically feasible and practically valuable for robotics. The ability to generate meshes on-demand opens new possibilities for scene understanding and sim-to-real transfer, while our integration of graphics advances into robotics suggests further untapped opportunities from generative AI. As these technologies mature, we anticipate rapid 3D reconstruction becoming a fundamental capability for interactive robotic systems.





\bibliographystyle{plainnat}

\bibliography{refs}


\end{document}